\documentclass[runningheads]{llncs}
\usepackage[T1]{fontenc}
\usepackage{graphicx}
\usepackage{algorithm}
\usepackage{amsmath}
\usepackage[noend]{algpseudocode}
\usepackage{float}
\usepackage{stackengine}
\usepackage{multirow}
\usepackage{booktabs}
\usepackage{arydshln}
\usepackage{tikz}
\usepackage{amssymb}
\usepackage{subcaption}

\usepackage{xcolor}

\usepackage[hidelinks,colorlinks=true,linkcolor=blue,citecolor=blue]{hyperref}
\usepackage[capitalize]{cleveref}
\crefname{section}{Sec.}{Secs.}
\Crefname{section}{Section}{Sections}
\Crefname{table}{Table}{Tables}
\crefname{table}{Tab.}{Tabs.}

\begin{document}

\title{An Organism Starts with a Single Pix-Cell:     ~~~~~~~~A Neural Cellular Diffusion for High-Resolution Image Synthesis }

\titlerunning{GeCA}

\author{Marawan Elbatel\inst{1}
\and
Konstantinos Kamnitsas\inst{2,4,5}  \and Xiaomeng Li\inst{1,3}\thanks{Correspondence: \email{eexmli@ust.hk}} 
}

\authorrunning{M. Elbatel et al.} 

\institute{Department of Electronic and Computer Engineering, The Hong Kong University of Science and Technology, Hong Kong, China
\\ \email{{\{mkfmelbatel,eexmli\}@.ust.hk}}
 \and Department of Engineering Science, University of Oxford, Oxford, UK \\
 \email{konstantinos.kamnitsas@eng.ox.ac.uk}
 \and HKUST Shenzhen-Hong Kong Collaborative Innovation Research Institute, Shenzhen, China
 \and Department of Computing, Imperial College London, London, UK
 \and School of Computer Science, University of Birmingham, Birmingham, UK
}

\maketitle

%

\begin{abstract}
Generative modeling seeks to approximate the statistical properties of real data, enabling synthesis of new data that closely resembles the original distribution. Generative Adversarial Networks (GANs) and Denoising Diffusion Probabilistic Models (DDPMs) represent significant advancements in generative modeling, drawing inspiration from game theory and thermodynamics, respectively. Nevertheless, the exploration of generative modeling through the lens of biological evolution remains largely untapped. In this paper, we introduce a novel family of models termed Generative Cellular Automata (GeCA), inspired by the evolution of an organism from a single cell. GeCAs are evaluated as an effective augmentation tool for retinal disease classification across two imaging modalities: Fundus and Optical Coherence Tomography (OCT). In the context of OCT imaging, where data is scarce and the distribution of classes is inherently skewed, GeCA significantly boosts the performance of 11 different ophthalmological conditions, achieving a 12\% increase in the average F1 score compared to conventional baselines. GeCAs outperform both diffusion methods that incorporate UNet or state-of-the art variants with transformer-based denoising models, under similar parameter constraints. Code is available at:~\url{https://github.com/xmed-lab/GeCA}.

\end{abstract}

\begin{figure}[t]
    \centering    
\includegraphics[width=\textwidth]{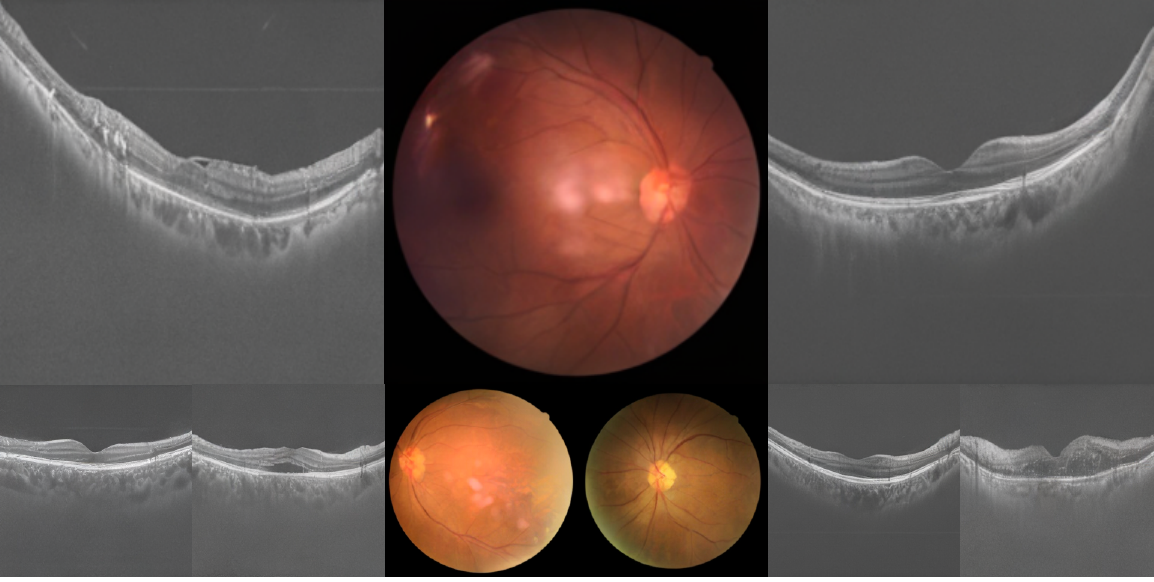}
    \caption{Selected synthetic images from our GeCA trained on Fundus and OCT.}    
\label{fig:intro}
\end{figure}

\section{Introduction}

Retinal diseases rank among the leading causes of vision disabilities and blindness if they remain untreated. Medical imaging modalities such as fundus photography and Optical Coherence Tomography (OCT) are widely used for diagnosing retinal conditions. OCT, offering a comprehensive view of the retinal layers compared to the fundus, is the preferred modality for diagnosing specific diseases such as Diabetic Retinopathy (DR) and Age-related Macular Degeneration (AMD)~\cite{midena2020optical}. Recently, deep learning approaches have been introduced for retinal disease screening, utilizing both fundus~\cite{nature_review_fundus_Li2024} and OCT~\cite{wang2023fundus}. Nevertheless, the development of these approaches is significantly hindered by the scarcity of publicly accessible datasets, particularly for OCT. Despite its advantages, OCT imaging is more costly and less employed than fundus photography, leading to a scarcity of OCT datasets. Therefore, it becomes crucial to develop a novel solution for retinal disease diagnosis using OCT imaging, especially considering its scarcity as well as its skewed disease distribution.

Expanding datasets with synthetic images through generative modeling has been shown to significantly enhance diagnostic accuracy in medical imaging, particularly in scenarios where data is scarce and class distribution is skewed~\cite{catasyn_miccai2023,miccai23_diffmic_cls,miccai23_dif_cls_seg_imb}. 
Current generative models primarily utilize diffusion-based optimization~\cite{DDPM}, relying heavily on architectures such as UNet~\cite{rombach2021highresolution_LDM,catasyn_miccai2023} and transformers~\cite{miccai23_dif_seg_transf,DiT_iccv_2023}. 
Despite their effectiveness, these models require a vast number of parameters, training on large-scale datasets, and often segmentation priors~\cite{Zhao2019SupervisedSO}. These inefficiencies present considerable challenges, particularly in medical imaging, where datasets, annotations, and computational resources are often scarce. Inspired by biological processes, Neural Cellular Automata (NCA)~\cite{mordvintsev2020growing} emerge as a promising alternative,  offering advancements in diverse tasks with fewer parameters~\cite{attention_nca,Pajouheshgar2022DyNCARD_dynca,kalkhof2023m3d,kalkhof2023med}. While NCA have shown promise in enabling medical image segmentation tasks under resource-constrained settings~\cite{kalkhof2023m3d,kalkhof2023med}, their application in generative tasks results in low-resolution outputs~\cite{palm2022variational_VAENCA,sudhakaran2022goalguided,Kalkhof2024FrequencyTimeDW_difnca} and lacks comprehensive performance comparisons, particularly in the evaluation of downstream tasks, where NCA's efficiency for image generation remains an unresolved challenge.

To address these challenges, we propose a novel approach for incorporating NCA in image generation by integrating diffusion objectives specifically devised for NCA's unique structure. 
Operating in the latent space, scaling Neural Cellular Automata (NCA) with transformers, and introducing a novel Gene Heredity guidance method for enhanced reverse sampling, we present Generative Cellular Automata (GeCA). GeCA surpasses the state-of-the-art Diffusion Transformers (DiTs)~\cite{DiT_iccv_2023} in image generation across two modalities: Fundus and OCT. By extending the application of GeCA to dataset expansion, we augment the scarce OCT dataset with synthetic images, resulting in a 12\% improvement in the average F1-score for multi-label retinal disease classification compared to conventional baselines. Our contributions can be summarized as:
\begin{itemize}
    \item We introduce {Generative Cellular Automata (GeCA)}, a novel model that integrates Neural Cellular Automata (NCA) with diffusion objectives, tailored specifically for NCA's unique structure.
    \item We propose Gene Heredity Guidance (GHG) to improve GeCA's image sampling.\textit{ GHG enabled GeCA to surpass SOTA DiT in image generation and retinal disease classification} with half of DiT's parameters.
        \item Through a detailed examination of diffusion models in OCT image generation, we demonstrate their capability to augment training datasets with synthetic images, boosting OCT's multi-label retinal disease classification.

\end{itemize}

\begin{figure}[t]
    \centering
    \includegraphics[width=\textwidth]{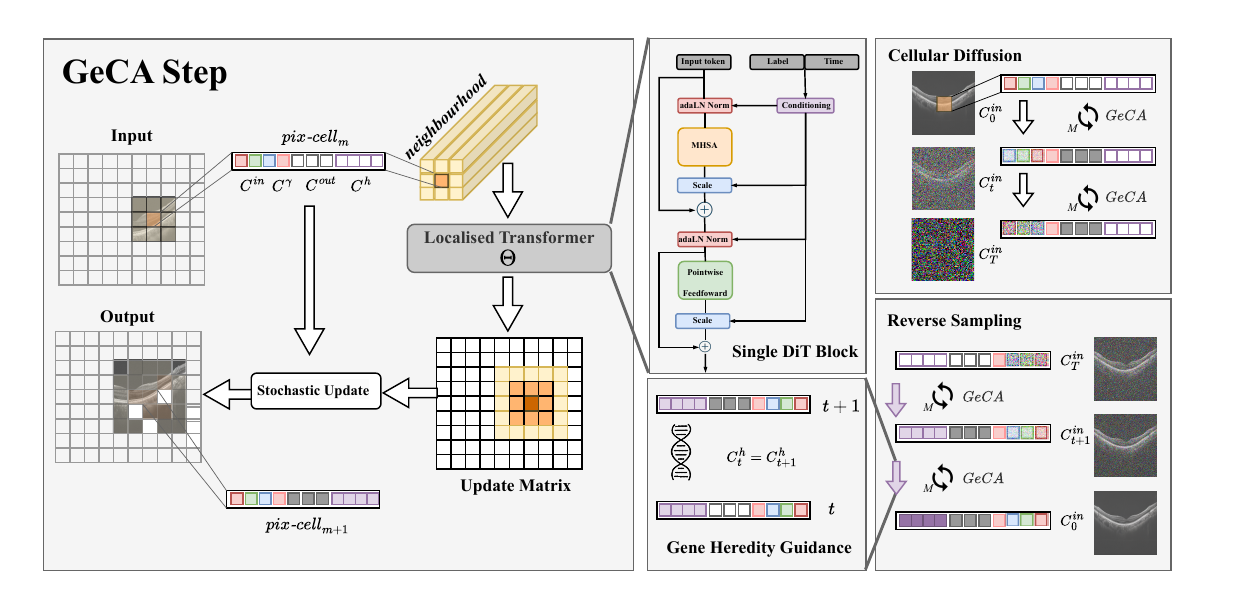}
    \caption{GeCA overall framework.}
    \label{fig:GeCA_framework}

\end{figure}

\section{Generative Cellular Automata}
\subsection{An Organism Starts With a Single Pix-Cell}

NCA~\cite{mordvintsev2020growing} model an input image with height $H$ and width $W$ as a grid comprising $H \times W$ entities, which we refer to as \emph{pix-cells} in our methodology. Each \emph{pix-cell} represents a time-dependent state space representation, facilitating dynamic evolution akin to cellular processes in an organism, \textit{i.e.}, image. We parameterize the state of each \emph{pix-cell} at step $m$ as a vector of scalars, defined as:

\begin{equation}
\text{\textit{pix-cell}}_{{m}} = \{C^{in}, C^{\gamma}, C^{out}, C^{h}\},
\end{equation}

where \(C^{in}\) denotes the input values of the image (e.g., one scalar for grayscale and three for RGB input images), $C^{\gamma}$ represents a positional encoding, defined by a continuous and smooth sinusoidal function facilitating spatial awareness within the grid~\cite{dosovitskiy2021an_vit,attention_is_all}, \(C^{out}\) refers to the output values indicating a \emph{pix-cells}'s targeted state (equivalent to \(C^{in}\) in dimension), and $C^{h}$ represents the hidden state variables reflecting the \textit{pix-cell}'s internal state during evolution.

To evolve a single \textit{pix-cell} to a more complex organism—an image, we follow traditional NCA conventional that adopts a stochastic rule~\cite{mordvintsev2020growing}. This means a \textit{pix-cell} is updated at step $m$ randomly with a probability $p$, reflecting the non-simultaneous nature of cellular updates in self-organizing organisms. The update of a \textit{pix-cell} focuses on updating only $C^{h}$ and $C^{out}$, given that $C^{in}$ and $C^{\gamma}$ are constant. This process, illustrated as GeCA step in~\cref{fig:GeCA_framework}, is defined as:

\begin{equation}
\textit{pix-cell}_{m+1}=\Theta(\textit{pix-cell}_{m}, \text{Neighborhood}_8) + \{0,0,C^{out}_{m}, C^{h}_{m}\}
\label{eq:update}
\end{equation}

Departing from the hierarchical modeling with $M$ layers in the SOTA Diffusion Transformer (DiT), we parameterize $\Theta$ as a \textit{single DiT block} featuring a localized self-attention mechanism, specifically computed across the 8 closest neighboring \textit{pix-cells}. The localized attention strategy, implemented similarly to those in localized transformer-based methods~\cite{zhang2021multi_longformer,chen2022regionvit,attention_nca}, allows each \textit{pix-cell} to grow independently by applying~\cref{eq:update} for $M$ times, using the same $\Theta$. GeCA's approach shifts the focus in image generation towards local spatial interactions, moving away from the global context reliance observed in traditional models such as UNet~\cite{UNet} and standard transformers~\cite{attention_is_all}. Nevertheless, GeCA achieves global coherence by accumulating long-term state-space representation via $C^{h}$, aligning with the foundational concepts documented in NCA~\cite{mordvintsev2020growing,sudhakaran2022goalguided,attention_nca,kalkhof2023med,kalkhof2023m3d,Pajouheshgar2022DyNCARD_dynca}, Mamba~\cite{Gu2023MambaLS},  universal transformers~\cite{dehghani2018universal_transformers}, and MLP-mixers~\cite{tolstikhin2021mlpmixer}.

\subsection{Cellular Diffusion: Evolving Cells into Organisms}

To train our model parameters $\Theta$, we utilize the well-established diffusion process first introduced in~\cite{DDPM} with specific modifications in the forward and reverse steps. During the forward diffusion process, we initialize $C^{out}$ and $C^h$ with zeros, except for a single \textit{pix-cell} located at the center of the $H \times W$ grid, which is initialized with random scalars to serve as the starting point for the cellular process. $C^{\gamma}$ is initialized with a sinusoidal positional encoding. $C^{in}$ can be described in the forward diffusion process on a per \textit{pix-cell} level as:

\begin{equation}
C^{in}_{t} = \sqrt{\alpha_t} C^{in}_{0} + \sqrt{1-\alpha_t} \epsilon, \quad \epsilon \sim \mathcal{N}(0, I),
\end{equation}
where $\epsilon$, following a normal distribution, represents the noise added at each step, and $\alpha_t$, which is part of a pre-defined variance schedule, takes values within the interval $(0, 1)$ for each time step $t=1$ to $T$.

We then
perform $M$ cellular updates with~\cref{eq:update} to developing $C^{out}_{t}$ and $C^{h}_{t}$. When $T \rightarrow \infty$, $C^{in}_{T}$ becomes equivalent to an isotropic Gaussian distribution~\cite{DDPM}. Thus, the optimization process is simplified from a theoretical formulation to predict the noise \(\epsilon\) from a \textit{pix-cell} as:

\begin{equation}
L = \mathbb{E}_{t \sim [1,T], C_{0,t}} \left[ \|\epsilon - C^{out}_t\|^2 \right]
\end{equation}
This formulation allows reverse sampling from a Gaussian noise $C^{in}_{T} \sim \mathcal{N}(0, \mathbf{I})$. Additionally, it allows adjusting $M$ during sampling to control the intensity of generation, from undergrowth to overgrowth; see Fig. \textcolor{blue}{6} in the appendix


\subsection{Improved Reverse Sampling via Gene Heredity}\label{sec:method_ghg}

Representing an input image with \textit{pix-cells}, a time-dependent state space representation, GeCA preserves long-term information within its internal hidden states, $C^{h}$, analogous to genetic material. Thus, we propose leveraging $C^{h}$ at time $t+1$ to guide the reverse generation of time $t$, mirroring the inheritance of genetic traits. Specifically, we modify each step in the reverse process to initiate the \textit{pix-cell} hidden states, $C^{h}$, as:

\begin{equation}
C^{h}_{t} =
\begin{cases} 
 \epsilon \sim \mathcal{N}(0, I) & \text{if } t = T \text{~and grid-center \textit{pix-cell}}, \\
C^{h}_{t+1} & \text{otherwise}.
\end{cases}
\end{equation}
Simultaneously, $C^{out}$, for the grid-center \textit{pix-cell} at each timestep is defined as:

\begin{equation}
C^{out}_{t}\sim \mathcal{N}(0, I)
\end{equation}

Our proposed process, termed \textit{Gene Heredity Guidance (GHG)}, sets the stage for denoising $C^{in}_t$ and refining $C^{h}_t$ from a plausible starting point. Following GHG, the denoising process to sample a synthetic \textit{pix-cell}, $C^{in}_0$,  adheres to traditional diffusion steps till $t \rightarrow 0$ as:

\begin{equation}
C^{in}_{t-1} = \frac{1}{\sqrt{\alpha_t}} \left(C^{in}_t - \frac{1-\alpha_t}{\sqrt{1-\alpha_{t-1}}} C^{out}_t\right),
\end{equation}

Note that without our proposed \textit{GHG}, the application of NCA in generative modeling is suboptimal (See~\cref{fig:GHG_ablation_study}).
 \subsection{Retinal Disease Classification}~\label{sec:method_retina}
Classifying retinal disease from OCT images presents significant challenges due to data scarcity and skewed class distributions. In light of these challenges, we leverage generative modeling to augment the dataset effectively, a strategy proven to significantly enhance downstream classification tasks compared to conventional augmentation techniques~\cite{catasyn_miccai2023,zhang2023expanding}. 

Following~\cite{zhang2023expanding}, we synthesize a training set expanded five-fold, mirroring the original training set's distribution. Given the original dataset's class distribution \(p_{\text{orig}}(y)\), with \(y\) representing the dataset labels and \(N_{\text{orig}}\) as the original dataset size, the objective is to expand the dataset five-fold to \(N_{\text{aug}} = 5 \times N_{\text{orig}}\), while preserving \(p_{\text{orig}}(y)\). This is achieved by ensuring that the count of each label \(y\) in the augmented dataset, \(\text{Count}_{\text{aug}}(y)\), is five times its original count as:

\begin{equation}
p_{\text{aug}}(y) = p_{\text{orig}}(y), \quad \text{where} \quad \text{Count}_{\text{aug}}(y) = 5 \times \text{Count}_{\text{orig}}(y)
\end{equation}
By preserving the original label distribution \(p_{\text{orig}}(y)\) in the augmented dataset, we maintain the dataset's inherent distribution to avoid any potential bias.

\begin{figure}[t]
    \centering
    \includegraphics[width=\textwidth]{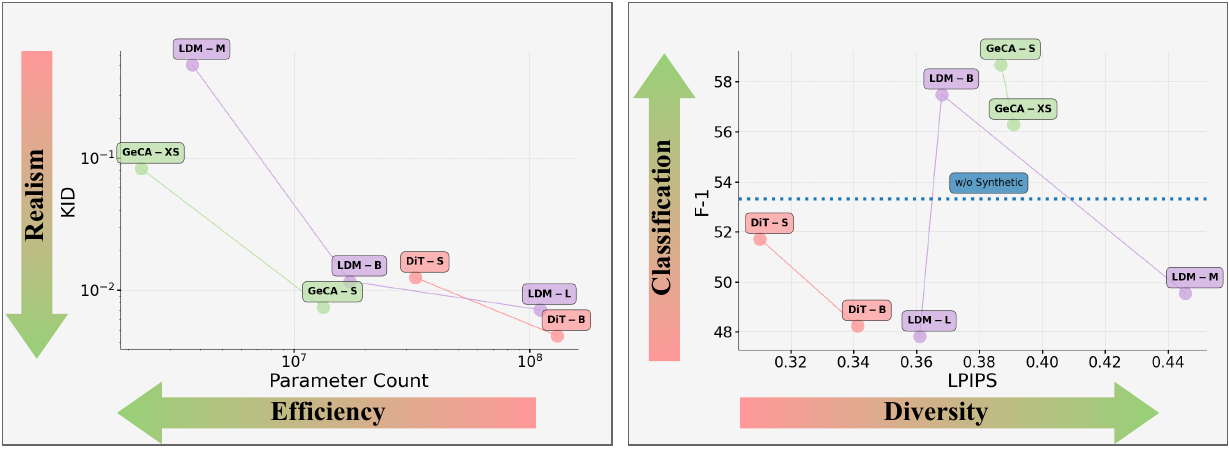}
    \caption{Results summary on a public fundus dataset.}
    \label{fig:fundus_results_summary}
\end{figure}

\section{Experiments}

\noindent{\textbf{Datasets.}} We evaluate our model using two different datasets: OCT and Fundus. The multi-label OCT dataset, {OCT-ML}, is an in-house dataset consisting of 1435 samples from 369 eyes of 203 patients considering multiple diseases including normal, dry age-related macular degeneration (dAMD), wet age-related macular degeneration (wAMD), diabetic retinopathy (DR), central serous chorioretinopathy (CSC), pigment epithelial detachment (PED), macular epiretinal membrane (MEM), fluid (FLD), exudation (EXU), choroid neovascularization (CNV), and retinal vascular occlusion (RVO). Additionally, we provide the code necessary for both the generation process and the classification task, applied to DeepDRiD~\cite{LIU2022100512_deepdrid},  \textit{a publicly available fundus imaging dataset} encompassing five grading classes 
and follow the MedMnist split~\cite{medmnistv2} (1,080 train, 120 val,  400 test). For the OCT-ML dataset, we adopt a five-fold cross-validation.

\begin{table}[h]
\centering
\caption{Quantitative image quality evaluation for two datasets. KID values are expressed in terms of \(10^{-3}\) for each model. All baselines are trained and evaluated with \textit{classifier free guidance (CFG)}~\cite{ho2021classifierfree} and $T$ = 250.}
\label{tbl:generative_results}
\resizebox{\textwidth}{!}{%
\begin{tabular}{l|c|ccc|ccc}
\hline
\multirow{2}{*}{\textbf{Method}} & \multirow{2}{*}{\textbf{\# Param. (\(\downarrow\)) }} & \multicolumn{3}{c|}{\textbf{Fundus Dataset}}                                   & \multicolumn{3}{c}{\textbf{OCT Dataset}}                                   \\ 
\cline{3-8} 
& & \textbf{{KID}} (\(\downarrow\)) & \textbf{{LPIPS}}~(\(\uparrow\))  & \textbf{GG}~($>0$)   & \textbf{{KID}} (\(\downarrow\))  & \textbf{{LPIPS}}~(\(\uparrow\)) & \textbf{GG}~($>0$)     \\

\hline

LDM-B~\cite{rombach2021highresolution_LDM} & 17.3 M & $11.64_{\pm 2.1}$ & $0.37_{\pm 0.09}$ & $-10.67$ & $64.5_{\pm 10}$ & $0.39_{\pm 0.16}$ & -2.31\\
DiT-S~\cite{DiT_iccv_2023} & 32.7 M & $12.45_{\pm 2.8}$ & $0.31_{\pm 0.09}$ & $-14.55$ & $62.3_{\pm 5.9}$ & $0.37_{\pm 0.14}$ & -0.44\\
\textbf{GeCA-S (ours)} & \textbf{13.3 M} & $\textbf{7.42}_{\pm \textbf{1.6}}$ & $\textbf{0.39}_{\pm \textbf{0.11}}$ & $\textbf{2.02}$ & $\textbf{49.1}_{\pm \textbf{8.0}}$ & $\textbf{0.53}_{\pm \textbf{0.16}}$ & \textbf{0.34}\\

\bottomrule
\end{tabular}
}
\end{table}

\noindent{\textbf{Baselines.}} Compared to previous NCA approaches~\cite{palm2022variational_VAENCA,Kalkhof2024FrequencyTimeDW_difnca} which exhibited suboptimal performance and did not compare with SOTA generative benchmarks, we compare our~\textit{GeCA} against DiT~\cite{DiT_iccv_2023}, \textit{state-of-the-art diffusion transformers}, as well as the U-Net-based diffusion models from LDM~\cite{rombach2021highresolution_LDM}, modifying the label embedding to support multi-label OCT generation. Training and inference for all baseline models adhere to the same hyperparameters with Classifier Free Guidance (CFG)~\cite{ho2021classifierfree} to facilitate conditional generation on downstream tasks. For the DiT, we report  DiT-S, with an optimal patch size of 2. Given our GeCA trains \textit{a single DiT layer}, we take $M=12$ equivalent to the number of layers in DiT-S; See Appendix for details.

\noindent{\textbf{Implementation Details.}} For all methods, generation is conducted in the latent space akin to LDM~\cite{rombach2021highresolution_LDM} with an output resolution of 256x256. Training acceleration for all methods is done with mixed-precision. We utilize a batch size of 128 and train all models for 14,000 epochs until convergence. 
For the downstream classification task, ResNet-34 is utilized with Adam optimizer.

\begin{figure}[t]
    \centering
    \includegraphics[width=\textwidth]{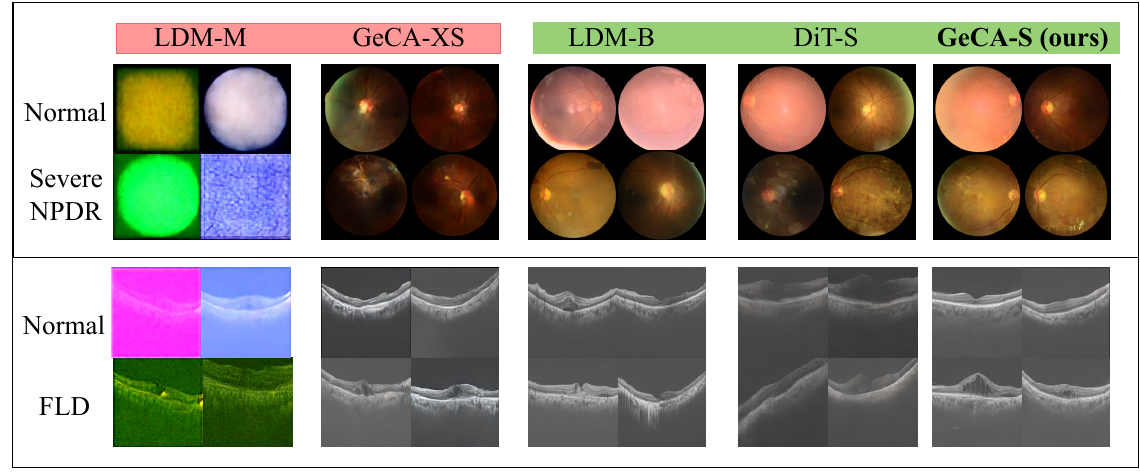}
    \caption{Qualitative examples for the Fundus and OCT datasets are provided; images are downsampled for visualization purposes, with high-resolution versions available in the supplementary material.}
    \label{fig:qualitative_examples}
\end{figure}

\begin{table}[t]
\centering
\caption{Performance results on our in-house multi-label OCT dataset, employing a five-fold cross-validation approach at the patient level. Each fold trains a separate diffusion model to generate synthetic data for the downstream classification task. All downstream experiments use ResNet34
as the backbone. We follow prior works~\cite{wang2023fundus,multi_modal_2} for eye-level performance evaluation, considering the multiple scans per eye in our dataset. The {F$1_{sen/pe}$} quantifies the harmonic mean of Sensitivity (Sen.) and Specificity (Spe.). ({****}) denote statistical significance with a p-value less than 0.0001. All reported metrics are macro-averaged.}
\resizebox{\textwidth}{!}{%
\label{tbl:oct_classification_results}
\begin{tabular}{l|ccccc|cc}
\hline
\textbf{Synthetic Data} & \textbf{Sen.} & \textbf{Spe.} & \textbf{AUC} & \textbf{F1} & \textbf{F$1_{sen/pe}$} & \textbf{mAP} & \textbf{$p<$} \\ \hline

Baseline (Geometric Aug) & 54.66$_{\pm 1.53}$ & \textbf{96.50$_{\pm 0.16}$} & 92.47$_{\pm 0.85}$ & 55.47$_{\pm 0.99}$ & 60.80$_{\pm 1.49}$ & 68.85$_{\pm 1.41}$ & - \\ 
Baseline w/o Aug. & 48.34$_{\pm 1.45}$ & 96.39$_{\pm 0.20}$ & 89.99$_{\pm 0.82}$ & 54.56$_{\pm 1.77}$ & 50.07$_{\pm 0.89}$ & 64.58$_{\pm 1.19}$ & ** \\ 
\hline

LDM-B~\cite{rombach2021highresolution_LDM} & 58.83$_{\pm 1.90}$ & 96.12$_{\pm 0.29}$ & 91.22$_{\pm 0.74}$ & 59.65$_{\pm 3.19}$ & 67.74$_{\pm 2.97}$ & 70.49$_{\pm 2.64}$ & ** \\

DiT-S~\cite{DiT_iccv_2023} & 59.25$_{\pm 4.54}$ & 95.87$_{\pm 0.37}$ & 91.80$_{\pm 1.74}$ & 59.11$_{\pm 2.57}$ & 67.13$_{\pm 4.87}$ & 69.89$_{\pm 3.34}$ & *** \\

\textbf{GeCA-S (ours)} & \textbf{59.95$_{\pm 5.32}$ }& 96.38$_{\pm 0.40}$ & \textbf{92.74$_{\pm 2.21}$} & \textbf{61.62$_{\pm 3.93}$} & \textbf{68.38$_{\pm 4.61}$} & \textbf{73.28$_{\pm 5.58}$} & **** \\

\hline
\end{tabular}
}
\end{table}

\begin{figure}[h]
    \centering
    \begin{subfigure}[b]{0.45\textwidth}
        \centering
        \includegraphics[width=\textwidth]{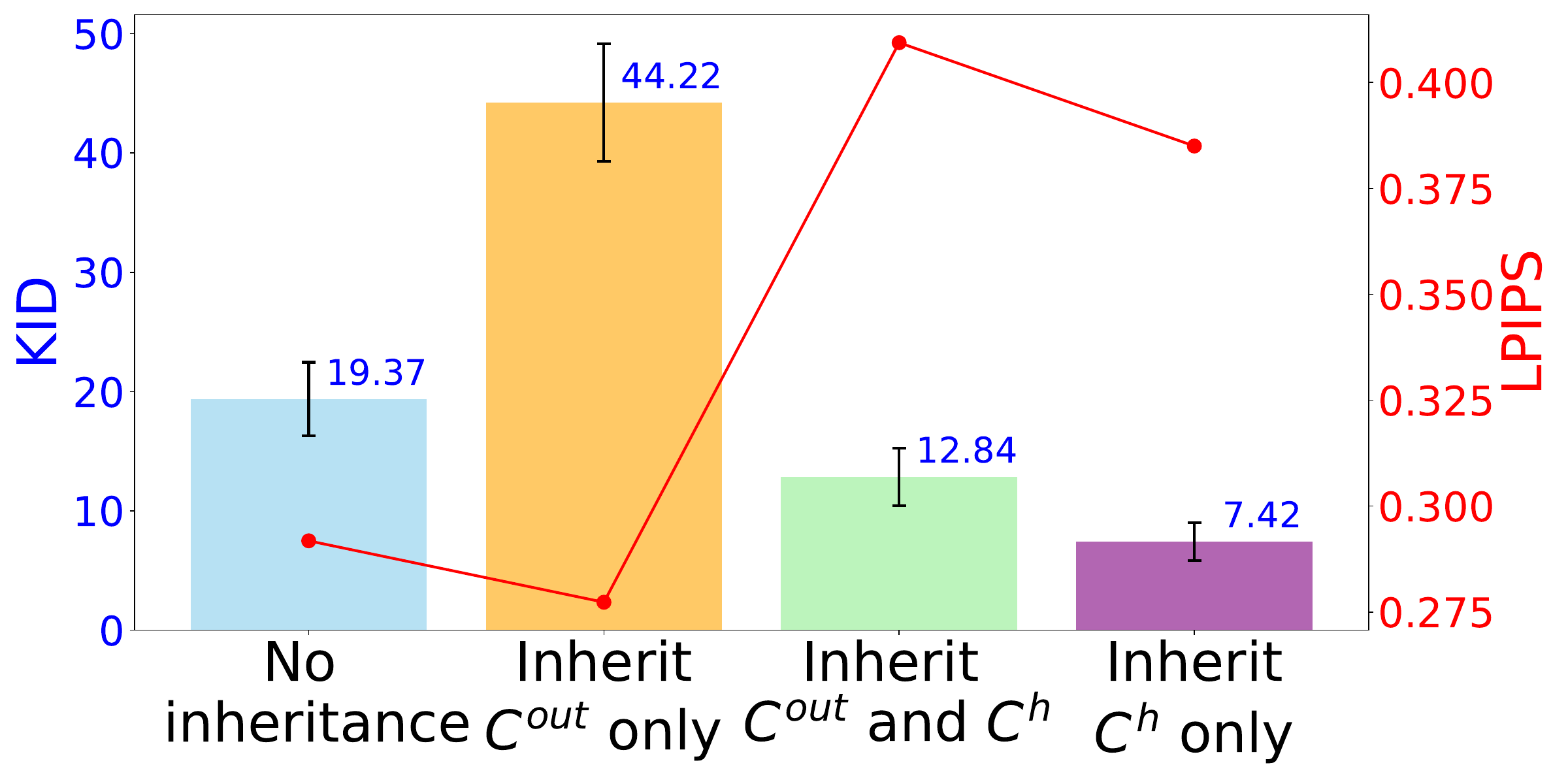}
        \caption{Public Fundus dataset.}
        \label{fig:sub1}
    \end{subfigure}
    \hfill
    \begin{subfigure}[b]{0.45\textwidth}
        \centering
        \includegraphics[width=\textwidth]{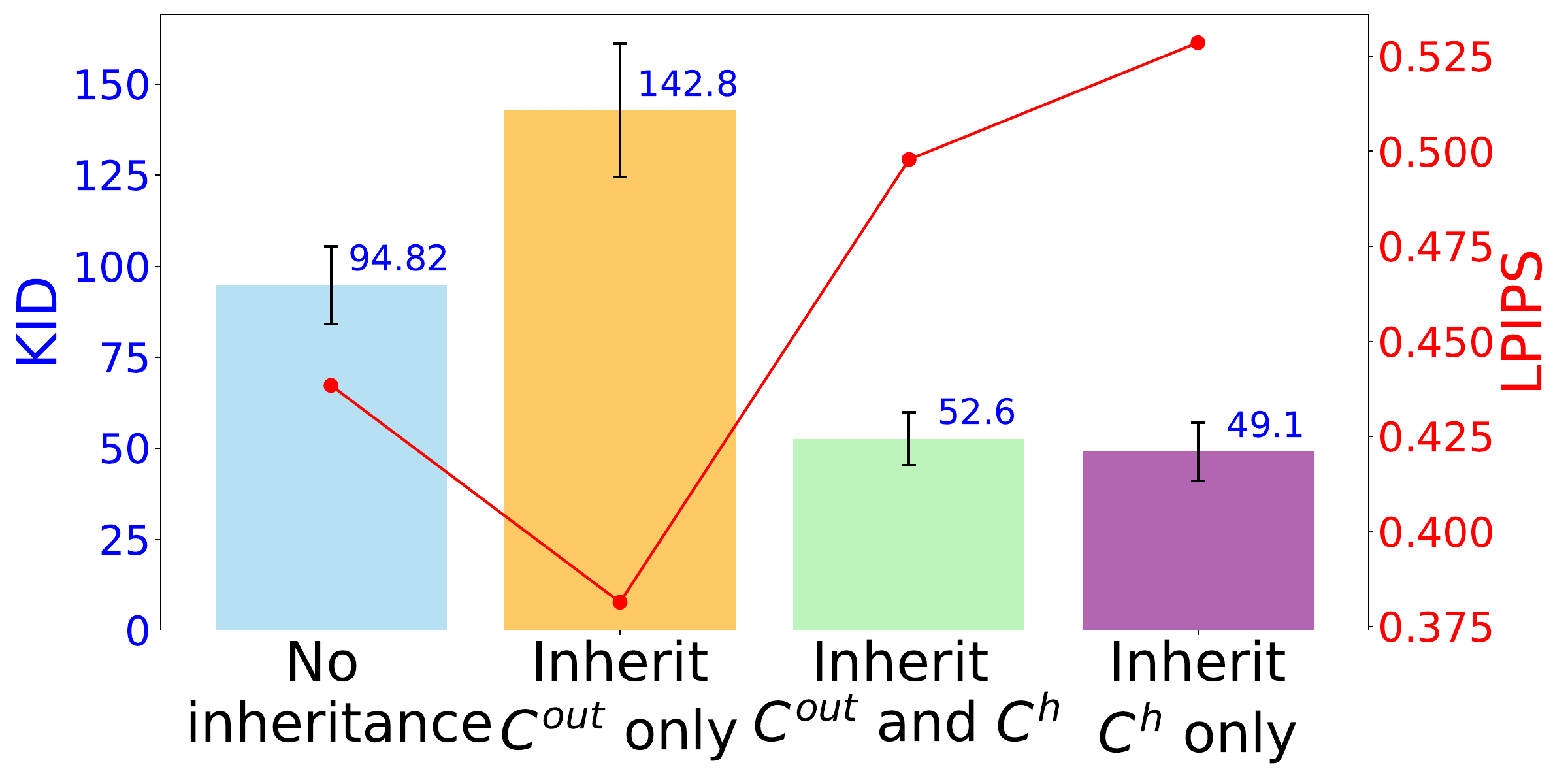}
        \caption{OCT-ML.}
        \label{fig:sub2}
    \end{subfigure}
    \caption{Ablation of the proposed Gene Heredity Guidance (GHG).}
    \label{fig:GHG_ablation_study}
\end{figure}

\noindent{\textbf{Generative Modeling Evaluation.}}~\cref{tbl:generative_results} presents the quantitative results to assess the quality of the generated samples. Noting the limitations of the Fréchet Inception Distance (FID) score observed in prior works~\cite{pmlr-v119-naeem20a_random_metric,jiralerspong2023feature_fld_google}, we employ the Kernel Inception Distance (KID) for~\textit{fidelity} due to its sensitivity to dataset size~\cite{jiralerspong2023feature_fld_google,bińkowski2018demystifying_kid}. Additionally, we report the the perceptual LPIPS \textit{diversity}~\cite{Zhang2018TheUE_LPIPS} to measure the image variability. Finally, we present the generalization gap (GG) as quantified by the Feature Likelihood Divergence (FLD)~\cite{jiralerspong2023feature_fld_google}, encapsulating the triplet \textit{novelty} (different from the training samples), \textit{fidelity}, and \textit{diversity} of the synthetic samples. Overall, our \textit{GeCA demonstrates superior image quality}, both quantitatively and qualitatively, as depicted in~\cref{fig:qualitative_examples}. We show samples from the high-resolution GeCA model in~\cref{fig:intro} and the appendix.

\noindent{\textbf{Retinal Disease Classification.}} \cref{tbl:oct_classification_results} presents the 11 multi-label classification results on \textit{OCT-ML} expanded by synthetic data via generative modeling discussed in~\cref{sec:method_retina}. All generative models remarkably improved the performance across various metrics. Notably, expanding the training dataset with our proposed~{GeCA} achieved the highest mean average precision (mAP of 73.28\%). {GeCA} significantly surpass the baseline with geometric augmentation by \textbf{4.43\%} in mAP and \textbf{7.58\%} in the harmonic mean of Sensitivity and Specificity ({F$1_{sen/pe}$}). Furthermore, in terms of the traditional F1-score, which evaluates precision and recall, {GeCA} achieved a significant \textbf{6.15\%} gain over the baseline. Despite being significantly more parameter-efficient, requiring only 40\% of the parameters compared to the SOTA DiT-S~\cite{DiT_iccv_2023}, GeCA still manages to surpass it by \textbf{3.39\%} in mAP. Furthermore, GeCA not only exceeds the performance of the leading baseline, LDM-B~\cite{rombach2021highresolution_LDM}, by \textbf{2.79\%} in mAP, but it also secures the highest degree of statistical significance ({****}). These results highlight GeCA's very promising performance in the realm of generative modeling.

\noindent{\textbf{GHG~Ablation.}}~\label{sec:ablation}
\cref{fig:GHG_ablation_study} reveals the impact of Gene Heredity Guidance (GHG), introduced in~\cref{sec:method_ghg}, on two datasets. On the Fundus dataset, without inheritance, the model yields a moderate KID of 19.37, lacking the benefits of long-range dependencies. Inheriting \( C^{out} \) alone drastically impairs performance, spiking the KID to 44.22, suggesting that inheriting \( C^{out} \) propagates noise. 
Conversely, inheriting both \( C^{out} \) and hidden states \( C^{h} \) partially mitigates this effect, reducing the KID to 12.84. Optimal performance is observed when only \( C^{h} \) is inherited, achieving the \textit{lowest KID of} 7.42. In contrast to \(C^{out}\), whose primary function is to predict noise, inheriting \(C^{h}\) facilitates the propagation of long-range dependencies, capturing the global context across the image.

\section{Conclusion}
We present GeCA, an innovative model outperforming current image generation benchmarks through neural cellular automata, demonstrated on challenging multi-label OCT classification. Future directions include broadening GeCA's validation across various domains and exploiting its unique capabilities, such as channel dimension selective sampling and temporal scheduling of its updates.


\begin{credits}
\subsubsection{\ackname} This work was supported in part by the grants from Foshan HKUST Projects, Grant Nos. FSUST21-HKUST10E and FSUST21-HKUST11E and in part by
Project of Hetao Shenzhen-Hong Kong Science and Technology Innovation Cooperation Zone (HZQB-KCZYB-2020083).
Marawan Elbatel is supported by the Hong Kong PhD Fellowship Scheme (HKPFS) from the Hong Kong Research Grants Council (RGC), and by the Belt and Road Initiative from the HKSAR Government.

\subsubsection{\discintname}
The authors have no competing interests to declare that are relevant to the content of this article.
\end{credits}

\bibliographystyle{splncs04}
\bibliography{main}
\clearpage

\title{Appendix for “GeCA”}

\appendix

\section*{\centering{Appendix for ``GeCA''}}

\setcounter{figure}{5}
\setcounter{table}{2}

\begin{figure}[h]
    \centering
\includegraphics[width=0.8\textwidth]{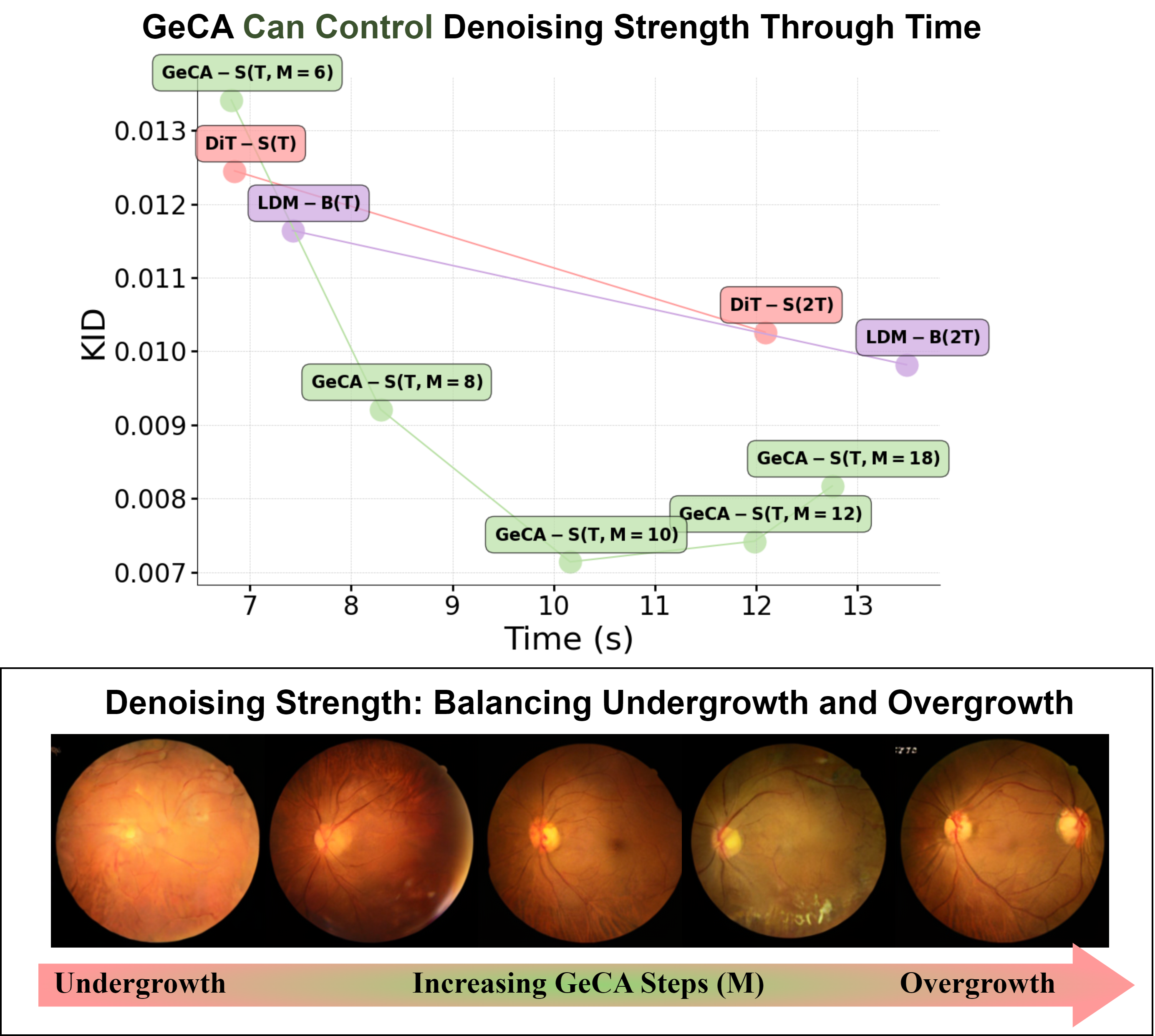}
\caption{Diverging from conventional hierarchical models, GeCA offers denoising strength adjustment via \textbf{M updates at each denoising step} without the need for re-training. Future efforts will investigate M-scheduling techniques. Performance is demonstrated under $T=250$. While speed concerns exist, GeCA's promising performance and optimization prospects highlight its significance.}

\label{app:fig:growth intensity}
\vspace{-1.4cm}
\end{figure}

\begin{table}[h]
\centering
\caption{Distribution of diseases across our in-house OCT images illustrating the uneven distribution of various ocular conditions within the dataset (class imbalance). All models trained on OCT images were subjected to rigorous validation using a 5-fold cross-validation process, with patient-wise splitting. Generative models are trained exclusively on the training set of each fold.}
\resizebox{0.9\textwidth}{!}{%
\begin{tabular}{l|c|c|c|c|c|c|c|c|c|c|c|c}
\hline
{Diagnosis} & {Normal} & {dAMD} & {wAMD} & {DR} & {CSC} & {PED} & {MEM} & {FLD} & {EXU} & {CNV} & {RVO} & {Total} \\ \hline
Count       & 278            & 160           & 145           & 502          & 95            & 133          & 196          & 613          & 573          & 138          & 34           & 1435          \\ \hline
\end{tabular}}
\label{app:tab:oct_images}
\end{table}

\begin{figure}[h]
    \centering
    \includegraphics[width=\textwidth]{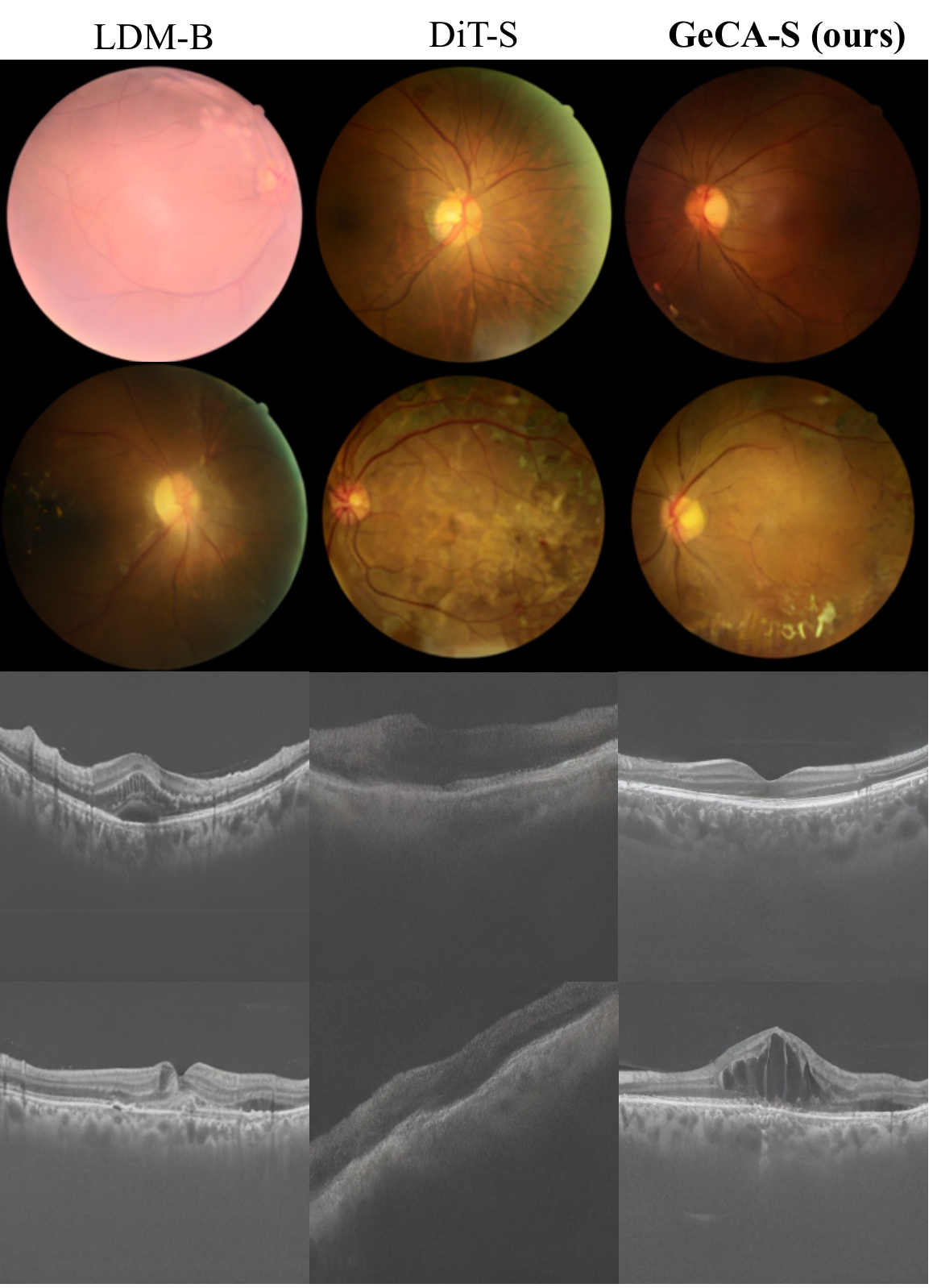}
    \caption{High Resolution output visualization.}
    \label{fig:enter-label}
\end{figure}

\end{document}